\documentclass{article}
\usepackage{spconf,amsmath,graphicx,cite}
\usepackage{amssymb}
\usepackage{multirow}
\usepackage{booktabs}
\usepackage[hidelinks]{hyperref}

\title{CiMRAG: CiM-aware Domain-adaptive and Noise-resilient Retrieval-Augmented Generation for Edge-based LLMs}

\name{Shih-Hsuan Chiu and Ming-Syan Chen}
\address{National Taiwan University, Taipei, Taiwan \\
\texttt{shchiu@arbor.ee.ntu.edu.tw, mschen@ntu.edu.tw}}

\begin{document}
\ninept
\maketitle

\begin{abstract}
Personalized virtual assistants powered by large language models (LLMs) on edge devices are attracting growing attention, with Retrieval-Augmented Generation (RAG) emerging as a key method for personalization by retrieving relevant profile data and generating tailored responses. However, deploying RAG on edge devices faces efficiency hurdles due to the rapid growth of profile data, such as user-LLM interactions and recent updates. While Computing-in-Memory (CiM) architectures mitigate this bottleneck by eliminating data movement between memory and processing units via in-situ operations, they are susceptible to environmental noise that can degrade retrieval precision. This poses a critical issue in dynamic, multi-domain edge-based scenarios (e.g., travel, medicine, and law) where both accuracy and adaptability are paramount. To address these challenges, we propose \textit{\underline{T}ask-\underline{O}riented \underline{N}oise-resilient \underline{E}mbedding \underline{L}earning} (TONEL), a framework that improves noise robustness and domain adaptability for RAG in noisy edge environments. TONEL employs a noise-aware projection model to learn task-specific embeddings compatible with CiM hardware constraints, enabling accurate retrieval under noisy conditions. Extensive experiments conducted on personalization benchmarks demonstrate the effectiveness and practicality of our methods relative to strong baselines, especially in task-specific noisy scenarios.
\end{abstract}
\vspace{0.17em}


\section{Introduction}
Large language models (LLMs) have become essential across many applications due to their strong reasoning capabilities \cite{minaee2024large, xi2025rise}. To meet the growing demands for personalized interactions, deploying these LLMs on edge devices (what we term "edge LLMs") has gained significant traction \cite{qu2025mobile, zheng2025review, wang2024comprehensive}. For effective personalization, edge LLMs must adapt to profile data such as user-LLM interaction or recent updates. However, as with cloud-based LLMs, edge LLMs primarily rely on fine-tuning model parameters, which is impractical for edge devices due to their limited computing power and memory \cite{qin2024empirical, xu2024device}.

\begin{figure}[t]
    \centering
    \includegraphics[width=1\linewidth]{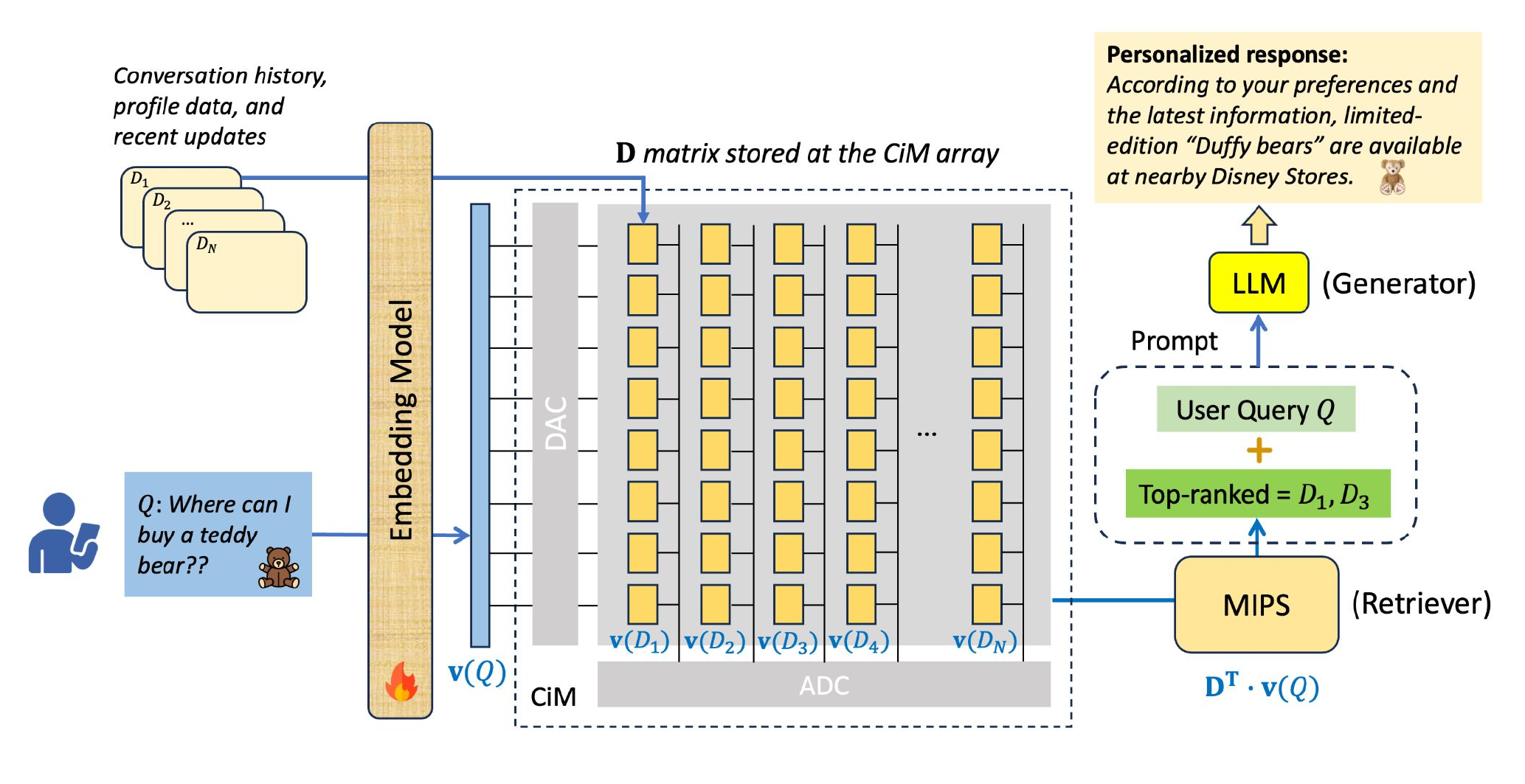}
    \caption{A schematic depiction of RAG for edge LLMs with CiM devices; the workflow is inspired by \cite{qin2024robust}. CiM employs MIPS to retrieve top-ranked documents, which are combined with the user query to enable the LLM to generate personalized responses. This paper focuses on enhancing the \textit{text embedding model} to produce \textit{\textbf{task-aware, noise-resilient CiM-friendly embeddings for MIPS}} in domain-specific edge environments.}
    \label{figure:CiM}
\end{figure}

Retrieval-Augmented Generation (RAG) has emerged as the de facto alternative, enabling personalization of edge LLMs without fine-tuning \cite{fan2024survey, lewis2020retrieval}. RAG retrieves the most semantically relevant documents to the input query via \textit{maximum inner product search (MIPS)} \cite{lewis2020retrieval} and feeds them, together with the query, into an LLM to generate personalized responses. Of note, all documents from profile data are first converted into numerical embedding vectors by an embedding model and stored as a matrix within an RAG system used for MIPS, as illustrated in Figure \ref{figure:CiM}.

Despite its efficiency on edge devices, RAG faces two key challenges for real-time interactions. First, as user data (e.g., conversation history with LLMs) grows beyond RAM, reliance on slower storage (e.g., HDDs or SSDs) increases data transfer latency \cite{kang2013enabling}. Second, the efficiency of MIPS-based retrieval degrades with larger datasets, making RAG impractical for extensive user data \cite{lewis2020retrieval}. Fortunately, a seminal work by Qin et al. \cite{qin2024robust} leveraged computing-in-memory (CiM) architectures \cite{banagozar2019cim} to accelerate RAG by speeding up matrix-vector multiplication, the core operation in MIPS. As shown in Figure \ref{figure:CiM}, CiM performs computations directly in memory, reducing data movement (i.e., between memory and processing units) and boosting efficiency \cite{sze2017efficient, peng2019dnn}.

\begin{figure*}[t]
    \centering
    \includegraphics[height=0.346\linewidth]{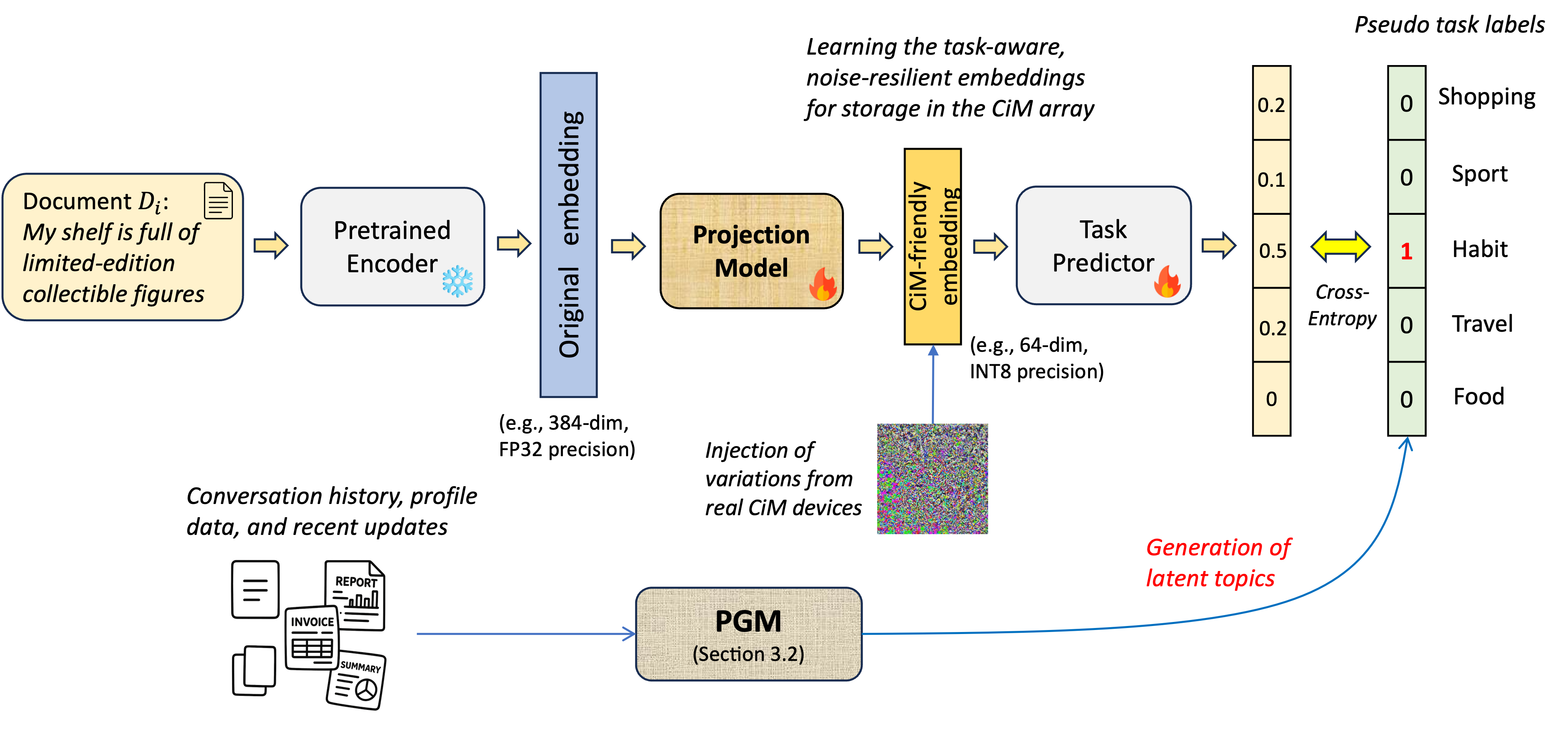}
    \caption{The overview of the proposed \texttt{TONEL} framework. It focuses on optimizing the \textit{projection model} to produce \textit{task-aware, noise-resilient CiM-friendly vectors} (in gold) suitable for MIPS in domain-specific edge environments.}
    \label{figure:TONEL}
\end{figure*}

Nevertheless, CiM arrays often rely on non-volatile memory (NVM), which is susceptible to environmental noise (e.g., temperature, humidity, and aging), potentially corrupting stored document embeddings and impairing retrieval (MIPS) performance \cite{yan2024compute}. To address this, Qin \cite{qin2024robust} proposes a noise-aware contrastive learning method that trains an embedding model to generate noise-resilient embeddings by pulling similar data closer and pushing dissimilar data apart, showing promising results. However, in practice, edge LLMs are deployed across diverse domains such as travel, medicine, and law. In such environments, environmental noise does not merely introduce random errors to the embeddings stored in CiM but significantly degrades the retrieval precision required for domain-specific tasks. Consequently, optimizing MIPS retrieval under noise in such dynamic, multi-domain scenarios remains a critical yet underexplored challenge. Moreover, real-world users often engage in multi-turn interactions with edge LLMs, leading to rapidly growing and topic-diverse profile documents, an issue not addressed in Qin's method \cite{qin2024robust}.

To address the above challenges, we identify two key issues: (i) \textit{MIPS robustness under noise}: Environmental noise (e.g., CiM variations) can distort matrix-vector similarity scores, necessitating a text embedding model that produces task-aware, noise-resilient embeddings. (ii) \textit{Label-free task adaptation}: In real-world scenarios, users interact with edge LLMs across diverse domains, producing a large volume of unlabeled documents. As manual labeling is impractical, the embedding model must adapt effectively in a label-free manner.

To this end, we propose \texttt{TONEL} (\textit{\underline{T}ask-\underline{O}riented \underline{N}oise-resilient \underline{E}mbedding \underline{L}earning}), a label-free framework that enables noise-robust and domain-adaptive MIPS retrieval on CiM hardware. TONEL incorporates two core components: \textbf{(1) To ensure MIPS robustness under noise}, we develop a \textit{noise-aware task-oriented optimization strategy (NATO)} that trains a projection model to generate task-specific embeddings compatible with CiM hardware. It uses pseudo topic labels under noisy conditions to produce noise-resilient representations. \textbf{(2) To enable label-free task adaptation}, we present a \textit{pseudo-label generation mechanism (PGM)} that guides the text embedding model to learn domain-adaptive, noise-resilient representations without manual annotation by generating latent task labels for NATO. Overall, TONEL is designed to \textit{enhance text embedding models to generate task-specific, noise-resilient embeddings tailored for MIPS on CiM hardware}.

\section{Problem Definition}
\subsection{RAG on CiM devices with MIPS-based retrieval}
An RAG system personalizes edge-based LLMs by retrieving relevant context from user profile data (viewed as a collection of documents). It comprises a text embedding model, a retriever, and a generator. Assume a user has $N$ documents (e.g., conversation history), each document $D_i$ is converted into a $d$-dimensional vector $\mathbf{v}(D_i)$ via embedding model $\mathrm{Emb(.)}$. These embeddings are assembled into a matrix $\mathbf{D}$ and stored in a CiM array, as shown in Figure \ref{figure:CiM}.
\begin{equation}
\mathbf{v}(D_i) = \mathrm{Emb}(D_i)\in \mathbb{R}^{d}, \quad \text{for } i = 1, \dots, N
\end{equation}
\begin{equation}
\mathbf{D} = [\mathbf{v}(D_1), \mathbf{v}(D_2), \dots, \mathbf{v}(D_N)] \in \mathbb{R}^{d \times N}
\end{equation}

During execution, the user query is also encoded into an embedding $\mathbf{v}(Q)$ using the same model. The retriever then performs \textit{maximum inner product search (MIPS)} \cite{lewis2020retrieval} by computing similarity scores via matrix-vector multiplication with stored document embeddings:
\begin{equation}
\mathbf{s} = \mathbf{D}^\top \mathbf{v}(Q) \in \mathbb{R}^{N}
\end{equation}

Finally, the RAG system concatenates the top-scoring (Top-K) retrieved documents with the query to form a prompt, which is then fed into the generator, typically an LLM, to produce a contextual response $\hat{Y}$ for the user:
\begin{equation}
\text{Prompt} = [D_i]_{i \in \mathrm{TopK}(\mathbf{s})} \, \Vert \, Q
\end{equation}
\begin{equation}
\hat{Y} = \mathrm{LLM}(\text{Prompt})
\end{equation}

Notably, document embeddings would be stored in a CiM array (Figure \ref{figure:CiM}, middle), which is vulnerable to environmental noise, potentially degrading retrieval (MIPS) accuracy. Thus, this work focuses on \textit{enhancing the text embedding model to produce embeddings that are robust to noise and adaptable across domains for MIPS in edge-based environments}.

\vspace{-0.2em}
\section{Proposed Method}
We propose \texttt{TONEL} (\textit{\underline{T}ask-\underline{O}riented \underline{N}oise-resilient \underline{E}mbedding \underline{L}earning}), a framework designed to enhance text embedding models that can generate task-aware, noise-resilient CiM-friendly embeddings for MIPS. As depicted in Figure \ref{figure:TONEL}, TONEL comprises two key components: (1) a noise-aware task-oriented optimization strategy (NATO), and (2) a pseudo-label generation mechanism (PGM).

\vspace{-0.1em}
\subsection{Noise-aware Task-oriented Optimization Strategy}
To tackle the challenge of MIPS robustness under noise, we develop a \textit{noise-aware task-oriented optimization strategy (NATO)} that trains a projection model to produce task-aware, noise-resilient embeddings while adhering to CiM hardware constraints. As shown in Figure \ref{figure:TONEL}, the training process first encodes all $N$ documents using a pretrained encoder $\mathrm{Enc(.)}$ (inherent to the LLM) to obtain embeddings with their original dimensionality and bit precision (e.g., 384-dimensional vectors in 32-bit floating-point format):
\begin{equation}
\mathbf{e}(D_i) = \mathrm{Enc}(D_i) \in \mathbb{R}^{384}_{\mathrm{FP32}}
\end{equation}

Since the document embeddings would be stored in a CiM architecture that is implemented as a "crossbar array" with typically fixed dimensions (e.g., 64$\times$64) and bit precision (e.g., 8-bit integers) \cite{qin2024robust, jiang2020device}, we employ a projection model $\mathrm{Proj(.)}$ to map them to 64 dimensions and apply simulated INT8 quantization via fake quantization and rounding \cite{jacob2018quantization}, generating CiM-friendly vectors that conform to crossbar array constraints:

\begin{equation}
\mathbf{e}(D_i)^\prime = \mathrm{Proj}(\mathbf{e}(D_i)) \in \mathbb{R}^{64}_{\mathrm{FP32}}
\end{equation}
\begin{equation}
\mathbf{e}(D_i)_q^\prime = \mathrm{clamp}\!\left( 
\left\lfloor \frac{\mathbf{e}(D_i)^\prime}{s} \right\rceil, 
-2^{b-1}, 2^{b-1}-1 \right)
\end{equation}
\begin{equation}
s = \frac{|max(\mathbf{e}(D_i)^\prime)|}{2^{b-1}-1}
\end{equation}
\begin{equation}
\tilde{\mathbf{e}}(D_i)= s \cdot \mathbf{e}(D_i)_q^\prime
\end{equation}

\noindent where $b=8$ (i.e., quantizing original vectors to 8-bit integers), $\left\lfloor \cdot \right\rceil$ denotes the round-to-nearest (RTN) operator, $\mathrm{clamp}(x,l,h)$ clamps $x$ to the range $[l,h]$, and the reconstruction vector $\tilde{\mathbf{e}}(D_i)$ is obtained using the uniform scaling factor $s$.

To account for hardware-induced perturbations, noise variations measured from real CiM devices are injected into the embeddings. The resulting noisy embeddings are fed into a task predictor $\mathrm{Pred(.)}$, which outputs a prediction score for each task. For model training, we employ the Cross-Entropy (CE) loss \cite{mao2023cross} and adapt it into the \textbf{\textit{CiM-aware Cross-Entropy (CiMCE) loss}} to incorporate the above factors, serving as our objective function. Given a set of documents $D = \{D_i\}_{i=1}^N = \{D_1, D_2 \ldots D_N\}$, each assigned to a class among $C$ classes, the CiMCE loss is formulated as:
\begin{equation}
\mathcal{L}_{\mathrm{CiMCE}} = -\frac{1}{N} \sum_{i=1}^{N} \sum_{c=1}^{C} \hat{y}_{i,c} \log P\big(c \mid \mathrm{Pred}(\tilde{\mathbf{e}}(D_i) + \eta)\big)
\end{equation}

Note that $\eta \sim \mathcal{N}(0, \sigma_v)$ refers to the CiM variation noise, modeled as Gaussian with zero mean and a standard deviation $\sigma_v$ specific to each value \cite{yan2022swim, yan2021uncertainty}. The $\hat{y}_{i,c}$ represents a pseudo task label, indicating whether document $D_i$ belongs to class $c$, generated by the PGM module (Section \ref{sec:PGM}). The model parameters are jointly optimized with task-specific information and noise perturbations, ensuring the embeddings remain discriminative under varying noise conditions. At inference time, the projection model produces task-aware, noise-resilient CiM-friendly embeddings (64 dimensions with 8-bit integers) for queries and documents, facilitating more accurate MIPS-based retrieval for domain-specific tasks.

\begin{figure}[t]
    \centering
    \includegraphics[width=1\linewidth]{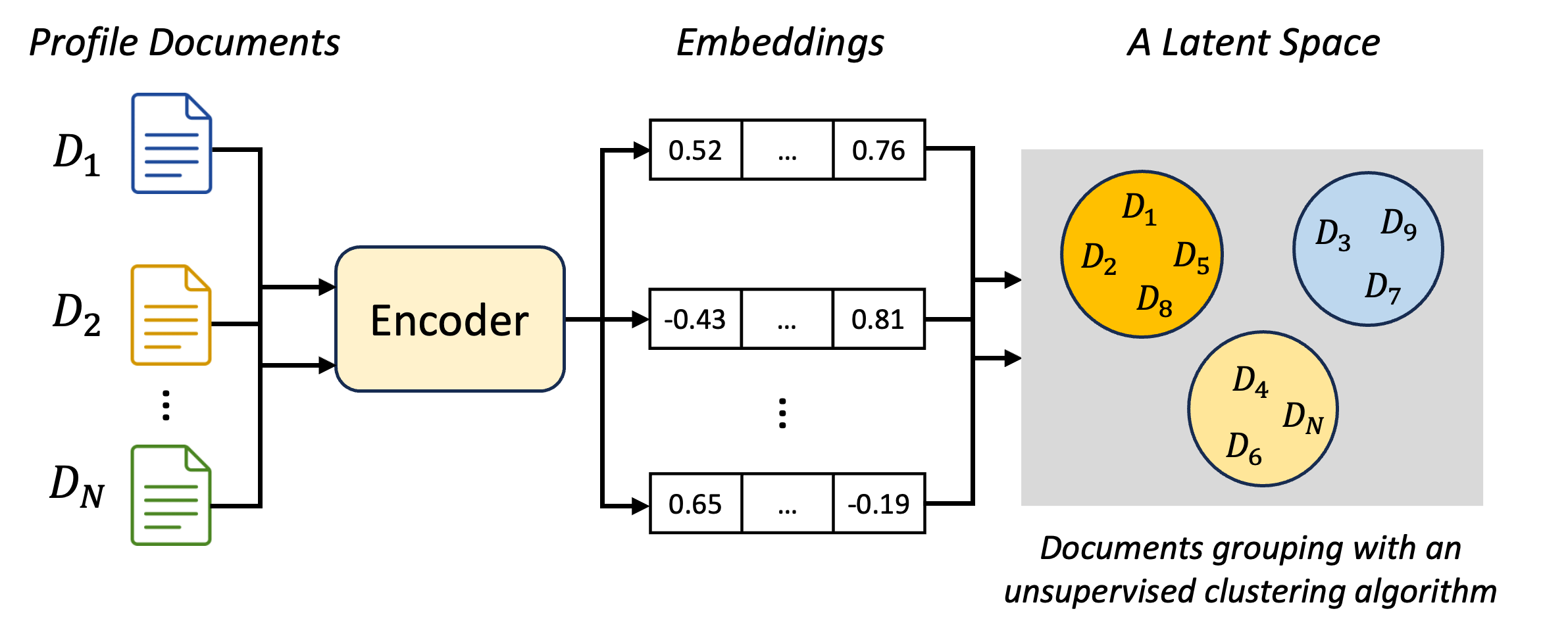}
    \caption{A schematic depiction of PGM for generating pseudo task labels used in TONEL.}
    \label{figure:PGM}
    \vspace{-0.2em}
\end{figure}

\vspace{-0.1em}
\subsection{Pseudo-Label Generation Mechanism}
\label{sec:PGM}
In real-world edge-based scenarios, such as travel or medical tasks, user-LLM interactions often span multiple topics and contexts over time, with rapidly growing profile data forming implicit task-level information. Incorporating these task-level cues into the embeddings of both documents and queries can significantly improve retrieval (i.e., MIPS) performance. However, manually annotating task-level information for each document is infeasible at scale. To flesh out this notion, we present a \textit{Pseudo-Label Generation Mechanism (PGM)} for TONEL (\textit{cf}. Figure \ref{figure:TONEL}) that supports its training process by alleviating the need for manually labeling the rapidly growing profile data with task-level information. More specifically, as depicted in Figure \ref{figure:PGM}, PGM assigns each document to one of the predefined $K$ groups by clustering their embeddings from the pretrained encoder $\mathrm{Enc(.)}$ using an unsupervised clustering algorithm (e.g., the $K$-means algorithm \cite{mcqueen1967some}). Each document $D_i$ is thus mapped to a specific group in a latent space, serving as its “pseudo” class $\hat{y}_{i}$. This PGM module enables TONEL to train the projection model with task-specific cues by automatically generating latent topics as pseudo task labels, thereby removing the need for manual annotation.

\begin{table}[t]
\centering
\caption{Device variations from different real CiM devices \cite{qin2024robust}.}
\vspace{1em}
\label{tab:device_variations}
\resizebox{0.918\columnwidth}{!}{%
\begin{tabular}{c c c c c}
\hline
\multirow{2}{*}{\textbf{Name}} & \multicolumn{4}{c}{\textbf{Device Variations $\sigma_v$}} \\
\cline{2-5} 
& $L_0$ & $L_1$ & $L_2$ & $L_3$ \\
\hline
\textit{RRAM}$_1$ (Device-1) & 0.0100 & 0.0100 & 0.0100 & 0.0100 \\
\textit{FeFET}$_2$ (Device-2) & 0.0067 & 0.0135 & 0.0135 & 0.0067 \\
\textit{FeFET}$_3$ (Device-3) & 0.0049 & 0.0146 & 0.0146 & 0.0049 \\
\textit{RRAM}$_4$ (Device-4) & 0.0038 & 0.0151 & 0.0151 & 0.0038 \\
\hline
\end{tabular}%
}
\vspace{0.15em}
\end{table}

\vspace{-0.3em}
\section{Empirical Experiments}
\vspace{-0.3em}
\subsection{Experimental Setup}
\noindent\textbf{Benchmark Datasets.} We evaluate TONEL on two personalization datasets from the LaMP benchmark: Movie Tagging (Movie) and Product Rating (Rating) \cite{salemi2023lamp}. In both datasets, each user has profile data consisting of textual history and corresponding labels. The Movie and Rating tasks are formulated as 15-class and 5-class classification problems, respectively.

\noindent\textbf{Comparative Methods.} We compare TONEL with two strong baselines (i.e., PCA and RoCR). PCA \cite{bishop2006pattern} is a classical, widely used dimensionality-reduction method, while RoCR \cite{qin2024robust} is a seminal approach that enables RAG on CiM architectures via noise-aware representation learning for documents.

\noindent\textbf{Evaluation Metrics.} Our primary metric is top-1 accuracy (Acc@1) of MIPS: the fraction of queries whose top-1 results from noisy CiM-friendly embeddings match the Oracle results from original full-precision embeddings without noise. We also present two metrics, Precision@5 (Prec@5), the ratio of oracle-relevant documents in the top-5, and nDCG@5, a position-aware score with respect to the oracle ranking. For downstream evaluation, we prepend the top-5 retrieved documents to the query and apply two representative edge-friendly LLMs (e.g., Gemma-2B \cite{team2024gemma} and Llama-3.2-3B \cite{dubey2024llama}) as generators, reporting classification Accuracy (Acc) and F1 score.

\noindent\textbf{Noise Settings.} Based on the device variations observed in real CiM devices shown in Table \ref{tab:device_variations}, the noise injection for document embeddings during testing follows \cite{qin2024robust, yan2021uncertainty}:
\begin{equation}
\operatorname{emb}(D_i)_{\sigma}^{d\times p}
= \bigl(e' \cdot L_0 \;+\; e' \cdot L_1 \;+\; e' \cdot L_2 \;+\; e' \cdot L_3 \bigr)\cdot \sigma
\end{equation}

\noindent where $e' = \operatorname{emb}(D_i)^{d \times p}$ is the embedding generated by the projection model, with $d$ and $p$ respectively denoting the reduced dimensionality and bit precision (e.g., $d = 64$, $p = 8$), and $\sigma$ representing the standard deviation of the injected Gaussian noise.

\begin{table}[t]
  \centering
  \caption{The MIPS (top-1 accuracy) results obtained by TONEL in comparison to that of baselines on the Movie and Rating datasets under noise variations from CiM Device-2 with different proportions of noisy documents.}
  \vspace{1em}
  \label{tab:on two datasets}
  \resizebox{0.735\columnwidth}{!}{
  \begin{tabular}{cc|cc}
    \toprule
    \textbf{Method} & \textbf{Noise (\%)} & \textbf{Movie} & \textbf{Rating} \\
    \midrule
    Oracle          & -     & 1       & 1       \\
    \midrule
    Random          & -     & 0.00068 &  0.00083   \\
    \midrule
    \multirow{3}{*}{PCA \cite{bishop2006pattern}}           
                    & Clean & 0.3478  &  0.0432     \\
                    & 50\%  & 0.3078  &  0.0383     \\
                    & 100\% & 0.2138  &  0.0346  \\
    \midrule
    \multirow{3}{*}{RoCR \cite{qin2024robust}}      
                    & Clean & 0.3826  &  0.0517  \\
                    & 50\%  & 0.3537  &  0.0485   \\
                    & 100\% & 0.3295  &  0.0453   \\
    \midrule
    \multirow{3}{*}{\textbf{TONEL (w/ PL)}} 
                    & Clean & 0.4313  &  0.0638   \\
                    & 50\%  & 0.4067 &  0.0566    \\
                    & 100\% & 0.3883  &  0.0584     \\
    \midrule
    \multirow{3}{*}{\textbf{TONEL (w/ TL)}} 
                    & Clean & 0.7667  &  0.2387  \\
                    & 50\%  & 0.7298  &  0.2336  \\
                    & 100\% & 0.7034  &  0.2452  \\
    \bottomrule
  \end{tabular}%
  }
  \vspace{-0.8em}
\end{table}

\vspace{-0.3em}
\subsection{Experimental Results}

In the first set of experiments, we compare TONEL against two strong baselines, PCA \cite{bishop2006pattern} and RoCR \cite{qin2024robust}, with Oracle and Random results listed for reference in Table \ref{tab:on two datasets}. Evaluations are conducted on two datasets (i.e., Movie and Rating) \cite{salemi2023lamp}. Notably, TONEL supports two training modes: the first leverages pseudo task labels from the PGM module (\textit{cf}. Section \ref{sec:PGM}), referred to as TONEL (w/ PL), and the second assumes access to ground-truth task labels, denoted as TONEL (w/ TL). To better assess our methods, we evaluate each approach under different proportions of noisy documents (Clean, 50\%, and 100\%). Two key observations can be drawn from Table \ref{tab:on two datasets}, which presents MIPS top-1 accuracy under noise variations from Device-2 (\textit{cf}. Table \ref{tab:device_variations}). First, our methods, TONEL (w/ PL) and TONEL (w/ TL), consistently outperform the baselines across both datasets and noise levels. Second, under 100\% noise (i.e., all document embeddings perturbed), TONEL (w/ PL) and TONEL (w/ TL) outperform RoCR by 6.2\% and 37.4\%, respectively, on the Movie dataset. These results indeed demonstrate the efficacy of TONEL to produce task-aware, noise-resilient embeddings tailored for MIPS.

In the second set of experiments, we evaluate our methods on the Movie dataset where all documents are perturbed (100\%) under noise variations induced by different real CiM devices (\textit{cf}. Table \ref{tab:device_variations}). We exhibit three evaluation metrics (Acc@1, Precision@5, and nDCG@5) to comprehensively assess the robustness and performance of TONEL methods in comparison to the baselines. The corresponding results are presented in Table \ref{tab:on all devices}. Inspection of Table \ref{tab:on all devices} reveals three noteworthy points. First, TONEL (w/ PL) consistently outperforms PCA and RoCR on all CiM devices and metrics, demonstrating its robustness to CiM-induced noise without requiring manual task labels. Second, TONEL (w/ PL) achieves average relative gains of 12.6\% in Precision@5 and 10.0\% in nDCG@5 over RoCR, indicating more accurate top-5 retrieval and ranking under noise conditions. Third, TONEL (w/ TL) delivers substantial improvements over all baselines and approaches Oracle performance, reflecting the full potential of TONEL with task supervision.

To further validate the practicality and feasibility of TONEL, we evaluate its downstream classification accuracy on the Movie Tagging dataset, a 15-category task (i.e., given a query of movie description, the LLM predicts a topic tag such as comedy and action). We select two representative edge-friendly medium-size LLMs (i.e., Gemma-2B \cite{team2024gemma} and Llama-3.2-3B \cite{dubey2024llama}) as generators to compare the performance of TONEL with Baseline and RoCR. Gemma-2B is one of the earliest open models by Google, with 4.95GB model weights. Llama-3.2-3B is a new SOTA open model by Meta, with 6.85GB model weights. The Baseline predicts a topic tag from the query alone, while RoCR and TONEL incorporate additional context by retrieving relevant profile documents. The corresponding results shown in Table \ref{tab:llm_comparison} yield three notable observations. First, augmenting the query with longer context (e.g., top-5 retrieved documents in this evaluation) via RAG improves LLM prediction accuracy and personalization compared to the Baseline, without modifying model parameters. Second, TONEL (w/ PL) outperforms RoCR, achieving relative improvements of 20.3\% in accuracy and 22.4\% in F1 score on Gemma-2B, confirming its effectiveness in a label-free setting. Third, TONEL (w/ TL) further improves over TONEL (w/ PL) by 26.1\% in accuracy and 41.0\% in F1 score on Llama-3.2-3B, reflecting its full potential when ground-truth task labels are available.

\begin{table}[t]
  \centering
  \caption{The Acc@1, Precision@5, nDCG@5 results obtained by TONEL in comparison to that of baselines on the Movie dataset under noise variations from different CiM devices.}
  \vspace{1em}
  \label{tab:on all devices}
  \resizebox{\columnwidth}{!}{%
  \begin{tabular}{cc|cccc}
    \toprule
    \textbf{Method} & \textbf{Metric} & \textbf{Device-1} & \textbf{Device-2} & \textbf{Device-3} & \textbf{Device-4} \\
    \midrule
    \multirow{1}{*}{Oracle} & - & 1 & 1 & 1 & 1 \\
    \midrule
    \multirow{3}{*}{PCA \cite{bishop2006pattern}}
      & Acc@1     & 0.2701 & 0.2138 & 0.1770 & 0.2092 \\
      & Prec@5    & 0.4839 & 0.4161 & 0.3621 & 0.4218 \\
      & nDCG@5    & 0.3833 & 0.3175 & 0.2748 & 0.3220 \\
    \midrule
    \multirow{3}{*}{RoCR \cite{qin2024robust}}
      & Acc@1     & 0.3531 & 0.3295 & 0.2713 & 0.3195 \\
      & Prec@5    & 0.6011 & 0.5736 & 0.5011 & 0.5218 \\
      & nDCG@5    & 0.5040 & 0.4718 & 0.3923 & 0.4291 \\
    \midrule
    \multirow{3}{*}{\textbf{TONEL (w/ PL)}}
      & Acc@1     & 0.3713 & 0.3883 & 0.3241 & 0.3368 \\
      & Prec@5    & 0.6471 & 0.6149 & 0.5977 & 0.6069 \\
      & nDCG@5    & 0.5154 & 0.4979 & 0.4731 & 0.4781 \\
    \midrule
    \multirow{3}{*}{\textbf{TONEL (w/ TL)}}
      & Acc@1     & 0.7701 & 0.7034 & 0.6885 & 0.6897 \\
      & Prec@5    & 0.8920 & 0.8701 & 0.8529 & 0.8678 \\
      & nDCG@5    & 0.8340 & 0.7920 & 0.7795 & 0.7856 \\
    \bottomrule
  \end{tabular}
  }
  \vspace{-0.5em}
\end{table}

\begin{table}[t]
  \centering
  \caption{The classification Accuracy and F1 results obtained by TONEL in comparison to that of baselines on the Movie dataset under noise variations from Device-2, using two edge-friendly LLMs (i.e., Gemma-2B and Llama-3.2-3B) as generators.}
  \vspace{1em}
  \label{tab:llm_comparison}
  \resizebox{0.82\columnwidth}{!}{%
  \begin{tabular}{c|cccc}
    \toprule
    \textbf{Method} & \multicolumn{2}{c}{\textbf{Gemma-2B}} & \multicolumn{2}{c}{\textbf{Llama-3.2-3B}} \\
    \cmidrule(lr){2-3} \cmidrule(lr){4-5}
                    & Acc ($\uparrow$) & F1 ($\uparrow$) & Acc ($\uparrow$) & F1 ($\uparrow$) \\
    \midrule
    Baseline        & 0.1460 & 0.0933 & 0.1084 & 0.0498 \\
    \midrule
    RoCR \cite{qin2024robust} & 0.3412 & 0.3107  & 0.3258 &  0.2893 \\
    \midrule
    \textbf{TONEL (w/ PL)} & 0.4104 & 0.3802  & 0.3974 & 0.3438 \\
    \midrule
    \textbf{TONEL (w/ TL)} & 0.5116 & 0.4780 & 0.5010 & 0.4847 \\
    \bottomrule
  \end{tabular}%
  }
  \vspace{-0.8em}
\end{table}

\vspace{-0.5em}
\section{Conclusion}
\vspace{-0.5em}
This work presents TONEL, a novel label-free framework that generates task-specific, noise-resilient embeddings compatible with CiM hardware for MIPS-based retrieval in RAG systems, enabling better personalization of LLMs in domain-specific edge environments. A series of empirical evaluations on personalization benchmarks demonstrates its practical utility over baselines and a representative RAG approach on CiM devices. Future directions include extending TONEL to better adapt to dynamic user profiles by developing efficient clustering algorithms co-designed with specialized hardware architectures, tailored for real-world edge-based LLM applications.

\vspace{-0.5em}
\section{Acknowledgment}
\vspace{-0.5em}
This research is supported in part by the National Science and Technology Council (NSTC), Taiwan, under Grant Number NSTC 114-2223-E-002-009, by the Ministry of Education, Taiwan, through the Higher Education Sprout Project—The Featured Area Research Center Program, and by the NTU–Delta Electronics Innovation Research Funding Project. Any findings and implications in the paper do not necessarily reflect those of the sponsors.

\bibliographystyle{unsrt}

\end{document}